\documentclass{article}

\usepackage{microtype}
\usepackage{graphicx}
\usepackage{subfigure}
\usepackage{booktabs} 

\usepackage{hyperref}

\usepackage[accepted]{icml2024}

\usepackage{amsmath}
\usepackage{amssymb}
\usepackage{mathtools}
\usepackage{amsthm}
\usepackage{makecell}
\usepackage{wrapfig}  
\usepackage{adjustbox}
\usepackage{mathrsfs}
\usepackage{listings}
\usepackage{pifont}
\usepackage{graphicx}
\usepackage{subfigure}
\usepackage{alltt}  
\usepackage{url}
\usepackage{multirow}
\usepackage{multicol}
\usepackage{float}  

\usepackage[capitalize,noabbrev]{cleveref}

\theoremstyle{plain}

\theoremstyle{definition}

\theoremstyle{remark}

\usepackage[textsize=tiny]{todonotes}

\icmltitlerunning{StrokeNUWA: Tokenizing Strokes for Vector Graphic Synthesis (Preprint)}
\begin{document}

\twocolumn[
\icmltitle{StrokeNUWA: Tokenizing Strokes for Vector Graphic Synthesis}



\icmlsetsymbol{equal}{*}

\begin{icmlauthorlist}
\icmlauthor{Zecheng Tang}{equal,1,comp}
\icmlauthor{Chenfei Wu}{equal,comp}
\icmlauthor{Zekai Zhang}{comp}
\icmlauthor{Mingheng Ni}{comp}
\icmlauthor{Shengming Yin}{comp}
\icmlauthor{Yu Liu}{comp}
\icmlauthor{Zhengyuan Yang}{comp2}
\icmlauthor{Lijuan Wang}{comp2}
\icmlauthor{Zicheng Liu}{comp2}
\icmlauthor{Juntao Li}{1}
\icmlauthor{Nan Duan}{comp}
\end{icmlauthorlist}

\icmlaffiliation{1}{Soochow University}
\icmlaffiliation{comp}{Microsoft Research Asia}
\icmlaffiliation{comp2}{Microsoft Azure AI}

\icmlcorrespondingauthor{Nan Duan}{nanduan@microsoft.com}

\icmlkeywords{Machine Learning, ICML}

\vskip 0.3in
]



\printAffiliationsAndNotice{\icmlEqualContribution} 

\begin{abstract}
To leverage LLMs for visual synthesis, traditional methods convert raster image information into discrete grid tokens through specialized visual modules, while disrupting the model’s ability to capture the true semantic representation of visual scenes. 
This paper posits that an alternative representation of images, vector graphics, can effectively surmount this limitation by enabling a more natural and semantically coherent segmentation of the image information.
Thus, we introduce StrokeNUWA, a pioneering work exploring a better visual representation \textemdash~``stroke tokens'' on vector graphics, which is inherently visual semantics rich, naturally compatible with LLMs, and highly compressed. 
Equipped with stroke tokens, StrokeNUWA can significantly surpass traditional LLM-based and optimization-based methods across various metrics in the vector graphic generation task. Besides, StrokeNUWA achieves up to a $94\times$ speedup in inference over the speed of prior methods with an exceptional SVG code compression ratio of 6.9\%.
\end{abstract}
\vspace{-2em}
\section{Introduction}
\label{sec:intro}

\begin{figure}[t]
    \centering
    \includegraphics[width=\columnwidth]{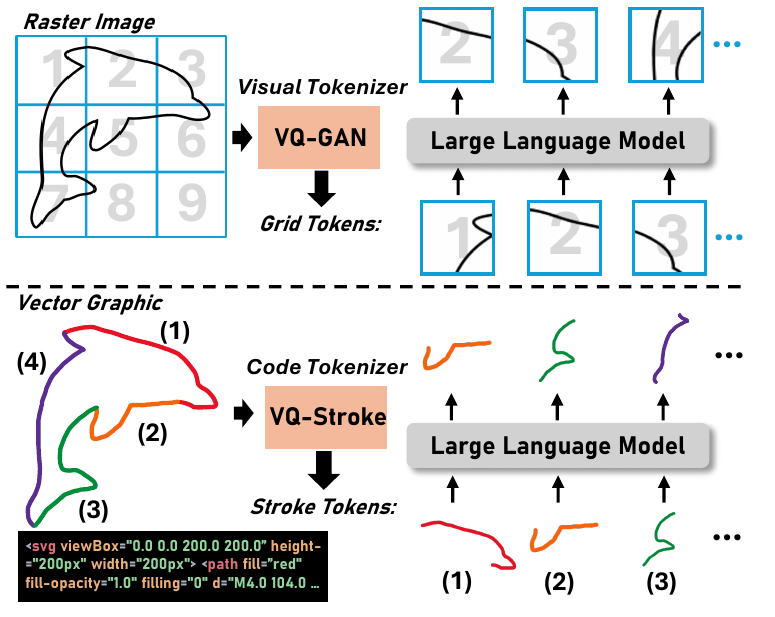}
    \vspace{-1em}
    \caption{Comparison between the visual representation of ``grid'' token  and our proposed ``stroke'' token. Instead of tokenizing pixels from raster images, we explore a novel visual representation by tokenizing codes, from another image format---Scalable Vector Graphic (SVG). ``Stroke'' tokens have the following advantages: (1) inherently contain visual semantics, (2) naturally compatible with LLMs, and (3) highly compressed.}
    \label{fig:1}
    \vspace{-1em}
\end{figure}

In recent years, Large transformer-based Language Models, commonly referred as LLMs, have made significant strides, particularly in the domain of Natural Language Processing~(NLP)~\cite{brown2020language,chowdhery2022palm,touvron2023llama,anil2023palm}. Concurrently, LLMs are gradually expanding their capabilities to other modalities, such as audio~\cite{ghosal2023text}, medical~\cite{singhal2023towards} and robotics~\cite{brohan2023rt}.

Current methodologies~\cite{reddy2021dall,wu2022nuwa,chang2022maskgit,kondratyuk2023videopoet} enable LLMs to generate visual information by transforming the continuous visual pixels into to discrete grid tokens via specialized visual modules such as VQ-VAE~\cite{van2017neural} and VQ-GAN~\cite{esser2021taming}. 
Subsequently, these transformed grid tokens are processed by the LLM in a manner akin to textual word handling, which facilitates LLMs' generative modeling process. However, when compared with diffusion models~\cite{rombach2022high}, LLMs still fall behind~\cite{lee2022draft,sun2023generative}.
The shortcomings of LLMs in visual tasks primarily arise from two reasons:
First, the transformation process relies on specific visual modules, which inherently possess limitations. For instance, advanced visual modules like VQ-GAN~\cite{esser2021taming} can lead to the generation of images with artifact~\cite{yu2023language};
Second, the use of grid tokens can disrupt the visual semantics, as the grids are artificially designed and not inherently semantic-aware. This artificial discretization imposes constraints on the model's ability to capture the true semantic representation of visual scenes.
\begin{figure*}[t]
    \centering
    \includegraphics[width=0.99\textwidth]{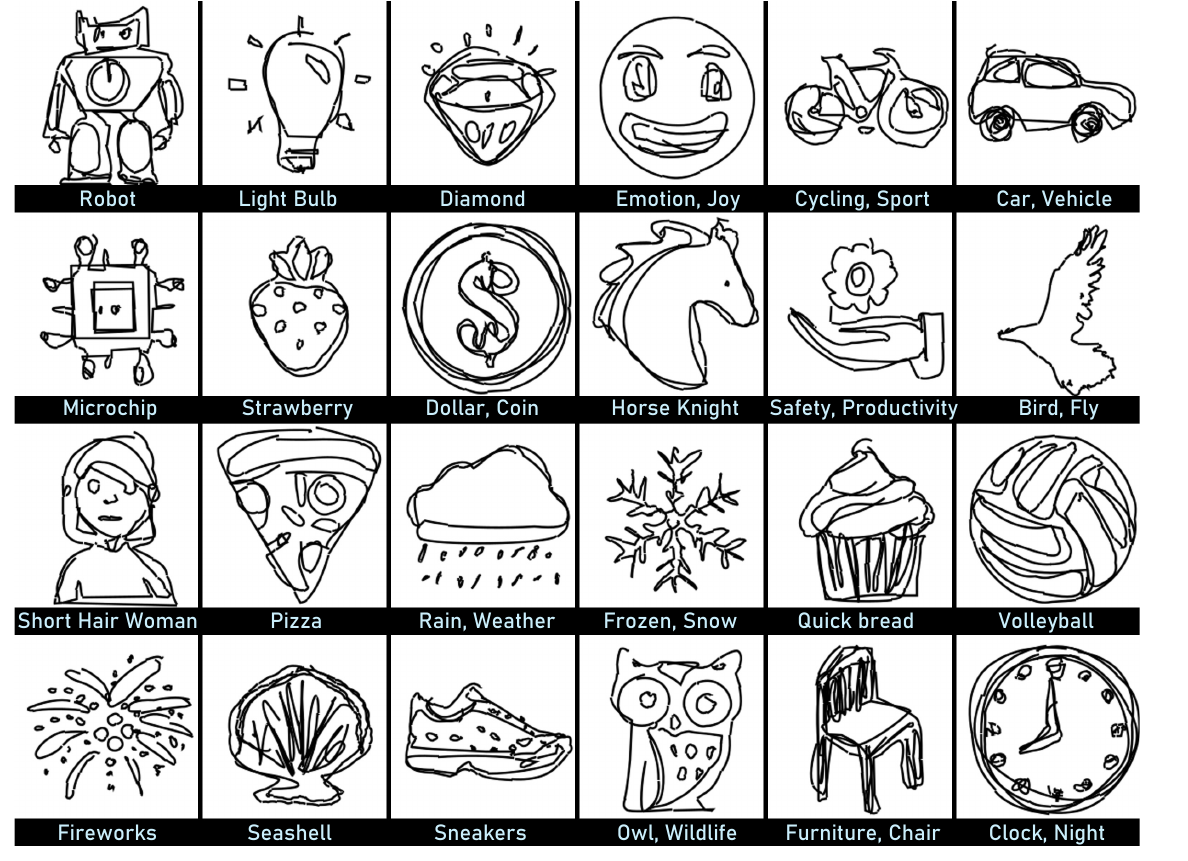}
    \vspace{-1em}
    \caption{SVG generated by StrokeNUWA. For each image, we provide partial keywords for clarity. }
    \vspace{-1em}
    \label{fig:enter-label}
\end{figure*}

\textit{Is there a visual representation that preserves the semantic integrity of visual information while being conducive to processing by LLMs?} 
Finding such a representation within the framework of grid tokens is non-trivial, as the arrangement of grid tokens is typically regular and uniform, whereas the semantic structure within images is often irregular and complex. As illustrated in Fig.~\ref{fig:1}, the dolphin's body is arbitrarily segmented into different grid tokens. Although there have been efforts to improve the VQ-VAE method~\cite{esser2021taming,yu2023language}, enhancing the visual representation quality, they are fundamentally constrained by the limitations inherent to raster image formats, leading to bottlenecks in semantic preservation.
In light of these challenges, we propose a novel approach that fundamentally retains the semantic concepts of images by utilizing an alternative image format: vector graphics. Different from pixel-based formats, vector graphics intrinsically reveal the construction of objects, naturally encapsulating the semantic concepts of the image. For example, our proposed ``stroke'' tokens segment the dolphin into sequentially connected strokes, where each stroke unit contains complete semantic information, such as the dolphin's fin~(stroke \ding{172}) and back~(stroke \ding{173}). 

It is worth mentioning that our intention is not to claim that vector graphics are superior to raster images, but rather to introduce a fresh perspective on visual representation. The advantages of our ``stroke'' token concept include: (1) Inherently contains visual semantics: each stroke token intrinsically contains visual semantics, offering a more intuitive semantic segmentation of the image content; (2) Naturally compatible with LLMs: the creation process of vector graphics is naturally sequential and interconnected, which mirrors the way LLMs process information. In other words, Each stroke is created in relation to the ones before and after it, establishing a contiguous and coherent sequence that LLMs can process more naturally; (3) Highly compressed: strokes in vector graphics can be highly compressed, allowing each stroke token to encapsulate a rich, compressed representation of the visual information, significantly reducing the data size while maintaining quality and semantic integrity.

Based on the above analysis, we introduce StrokeNUWA, a model that crafts vector graphics without the reliance on the visual module. StrokeNUWA consists of a VQ-Stroke module and an Encoder-Decoder model. The VQ-Stroke, based on the residual quantizer model architecture~\cite{martinez2014stacked}, can compress serialized vector graphic information into several SVG tokens. The Encoder-Decoder model primarily utilizes the capabilities of a pre-trained LLM to generate SVG tokens guided by text prompts.

We compare StrokeNUWA with optimization-based methods in the text-guided Scalable Vector Graphic~(SVG) generation task. Our approach achieves higher CLIPScore~\cite{hessel2021clipscore} metrics, suggesting that utilizing stroke tokens can yield content with richer visual semantics. When benchmarked against LLM-based baselines, our method surpasses them across all metrics, indicating that stroke tokens can integrate effectively with LLMs. Finally, due to the compression capabilities inherent in vector graphics, our model demonstrates significant efficiency in generation, achieving speed improvements of up to 94 times.

In a nutshell, our contributions can be outlined as follows:\vspace{-1em}
\begin{itemize} \itemsep -0.1em
    \item We introduce StrokeNUWA, the pioneering study exploring a better visual representation---stroke token, to synthesize vector graphics solely through LLMs without relying on specialized visual modules. 
    \item We propose VQ-Stroke, a specialized Vector Quantized Variational Autoencoder~(VQ-VAE) designed to compress vector graphics into stroke tokens, providing an exceptional compression ratio of 6.9\%.
    \item We conduct detailed experiments that demonstrate the significant potential of stroke tokens in the text-guided vector graphic synthesis task.
\end{itemize}
\section{Related Work}
\label{sec:background}

\subsection{Visual Representation}
\label{visual_synthesis_methods}
In the realm of computer graphics, two predominant image formats prevail: raster images, characterized by pixel matrices; and vector images, a.k.a, Scalable Vector Graphic~(SVG), characterized by a series of code language commands~\cite{zhang2023beyond}. Recent developments in visual synthesis have predominantly centered on the generation of raster images. The basic idea is to transform the continuous image pixels into discrete grid tokens via specialized visual modules such as VQ-VAE~\cite{van2017neural} and VQ-GAN~\cite{esser2021taming}, and then leverage LLMs to generate these tokens~\cite{reddy2021dall,wu2022nuwa,kondratyuk2023videopoet}. Most recently, some works have tried to improve ``grid'' tokens by designing advanced architectures such as Lookup-Free Quantization~\cite{yu2023language} and Efficient VQ-GAN~\cite{cao2023efficient}. However, these ``grid'' token representations can disrupt visual semantics as the grids are artificially designed, which lacks inherent semantic awareness, and are easily subject to the visual module's intrinsic limitations like disturbances and tampering~\cite{hu2023robust}.
Conversely, our study is a pioneering effort exploring a better visual representation by proposing the concept of the ``stroke'' token. Different from the `grid'' tokens, the ``stroke'' token is inherently defined by contextually associated coded language commands that offer strong semantic integrity, potentially mitigating the aforementioned issues.

\subsection{SVG Generation}
\label{subsec:svg_generation}
SVG generation employs a method of structured code generation for producing graphics, which offers better interpretability, flexibility, and scalability in image representation. 
The current mainstream approach of SVG generation is optimization-based methods~\cite{su2023marvel,jain2023vectorfusion,xing2023svgdreamer}, which share a similarity with traditional raster image generation, involving iteratively refining randomly initialized SVG paths to fit a target raster image with a differentiable rasterizer~\cite{li2020differentiable}.
However, the optimization process is both time-consuming and computationally intensive, e.g., creating an SVG graphic comprised of 24 SVG paths can exceed 20 minutes\footnote{We test with LIVE~\cite{ma2022towards} and VectorFusion~\cite{jain2023vectorfusion} on one NVIDIA V100 GPU.}.
Alternatively, some recent approaches have begun to adopt auto-regressive models to directly generate code for SVG synthesis~\cite{wang2022aesthetic,wu2023iconshop}.
However, due to the inherent extensive length nature of SVGs and a lack of effective SVG representation, these methods constrain LLMs to generate complex SVGs.
To address these challenges, we introduce VQ-Stroke and present the concept of ``stroke'' tokens. By transforming SVGs into stroke tokens, our approach enables LLMs to produce intricate SVGs with significantly improved inference speed.

\section{Methodology}
\begin{table}[t]
\centering
\small
\begin{tabular}{l c l c}
    \toprule
    \textbf{Name} & \textbf{Symbol} & \textbf{Argument} & \textbf{Example} \\
    \midrule
    \makecell[l]{Move \\ To} & \texttt{M} & \makecell[c]{($x_0$, $y_0$) \\ ($x_1$, $y_1$)} & 
    \begin{tabular}{c}
         \includegraphics[width=2.5cm]{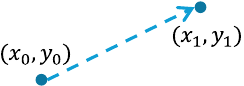}
    \end{tabular}
    \\
    \midrule
    \makecell[l]{Line \\ To} & \texttt{L} & \makecell[c]{($x_0$, $y_0$) \\ ($x_1$, $y_1$)}  & 
    \begin{tabular}{c}
         \includegraphics[width=2.5cm]{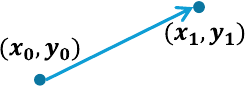}
    \end{tabular}
    \\
    \midrule
    \makecell[l]{Cubic \\ B\'{e}zier} & \texttt{C} & \makecell[c]{($x_0$, $y_0$) \\ ($x_1$, $y_1$) \\ ($c^{x}_0$, $c^{y}_0$) \\ ($c^{x}_1$, $c^{y}_1$)} &
    \begin{tabular}{c}
         \includegraphics[width=2.5cm]{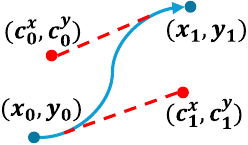}
    \end{tabular}
    \\
    \bottomrule
\end{tabular}
\caption{Overview of basic SVG commands, including \texttt{M}, \texttt{L}, and \texttt{C}, where each command contains one beginning point $(x_0, y_0)$ and one end point $(x_1, y_1)$. For Cubic B\'{e}zier command, it contains two extra control points $(c^{x}_0, c^{y}_0)$ and $(c^{x}_1, c^{y}_1)$.}
\vspace{-1em}
\label{tab:basic_commands}
\end{table}

\subsection{Problem Formulation}
\label{subsec:problem}
SVG code provides a suite of command and syntax rules, e.g., the ``\texttt{<rect>}'' command defines a rectangle shape with its position, width, and height, which can be written as \texttt{<rect x="10" y="20" width="50" height="80"/>}. 
However, considering the multitude of SVG command types, creating such a system not only requires a complex data structure, but without a massive dataset, LLMs would struggle to model the diverse range of commands effectively.
Therefore, as shown in Tab.~\ref{tab:basic_commands}, we can simplify each SVG using just three basic commands: ``Move To'', ``Line To'', and ``Cubic Bézier'' by following Iconshop~\cite{wu2023iconshop} and DeepSVG~\cite{carlier2020deepsvg}. For instance, intricate commands like ``\texttt{<rect>}'' can be constructed by those three basic commands.
After simplification, an SVG $\mathcal{G}=\{\mathcal{P}_i\}_{i=1}^N$ can be described with $N$ SVG paths, with each SVG path $\mathcal{P}_i$ consists of $M_{i}$ basic commands: $\mathcal{P}_{i}=\{\mathcal{C}_i^j\}_{j=1}^{M_i}$, where $\mathcal{C}_i^j$ is the $j$-th command in the $i$-th path.
Eventually, each basic command $\mathcal{C} =\left(T, \mathcal{V}\right)$ is consist of command type $T \in \{\texttt{M}, \texttt{L}, \texttt{C}\}$, and the corresponding position argument $\mathcal{V}$.

\vspace{-0.5em}
\subsection{StrokeNUWA}

StrokeNUWA contains three core components: a Vector Quantized-Stroke~(VQ-Stroke) for SVG compression, an Encoder-Decoder-based LLM~(EDM) for SVG generation, and an SVG Fixer~(SF) for post-processing. 
Firstly, VQ-Stroke compresses the SVG into stroke tokens, which enables a transformation between the SVG code and the discrete stroke tokens.
Then, EDM utilizes the stroke tokens produced from VQ-Stroke to generate SVG code. 
Finally, SF is a post-processing module designed to refine the quality of the generated SVGs, given that the output generated from the EDM or VQ-Stroke may not always conform to the stringent syntactical rules of SVG code.
Below, we will introduce the details of each component.

\begin{figure}[t]
    \centering
    \includegraphics[width=0.95\columnwidth]{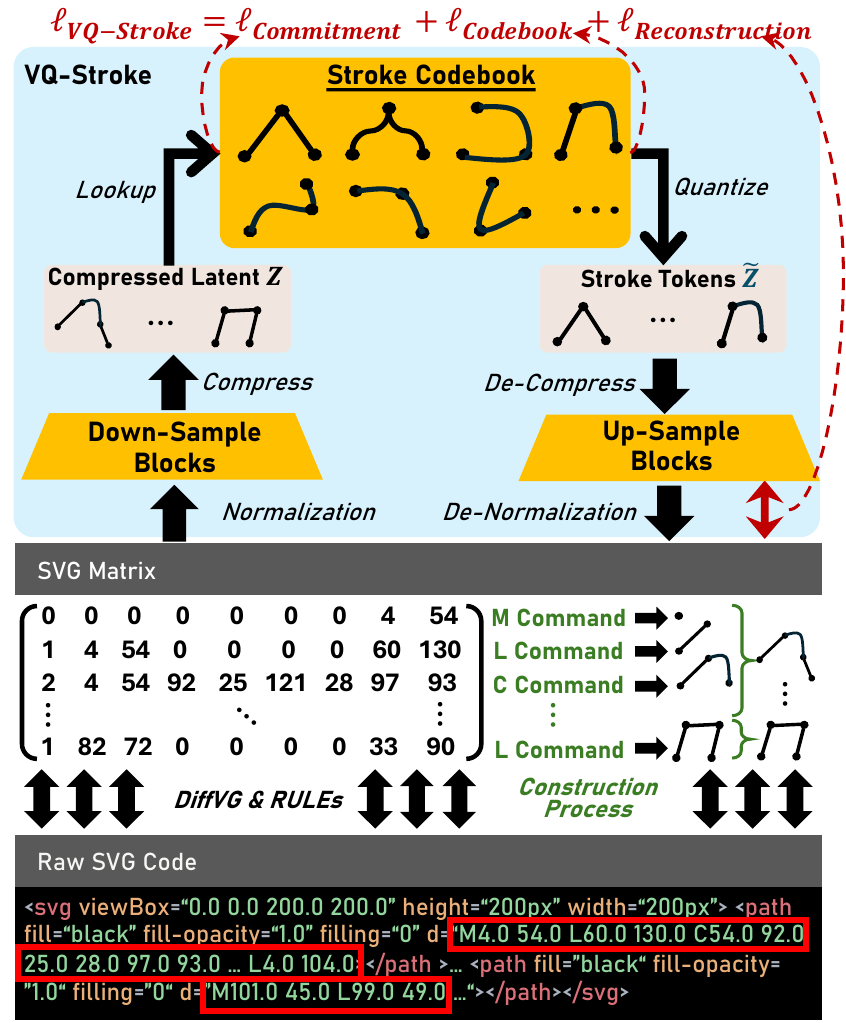}
    \vspace{-1em}
    \caption{Overview of VQ-Stroke.}
    \vspace{-1em}
    \label{fig:overview-VQ-Stroke}
\end{figure}

\subsubsection{Vector Quantized-Stroke}
\label{subsec_VQ-Stroke}

VQ-Stroke encompasses two main stages: ``Code to Matrix'' stage that transforms SVG code into the matrix format suitable for model input, and ``Matrix to Token'' stage that transforms the matrix data into stroke tokens.

\paragraph{Code to Matrix}

As depicted in Fig.~\ref{fig:overview-VQ-Stroke}, we first transform the simplified SVG code~(Sec.~\ref{subsec:problem}) into SVG matrix format by converting each basic command $\mathcal{C}_{i}^{j}$ to the individual vector $\mathcal{K}_{i}^{j} \in \mathbb{R}^{9}$ with rules $f$:
\begin{equation}
    \mathcal{K}_{i}^{j} = f(\mathcal{C}_{i}^{j}) = \left(T, x_0, y_0, c^{x}_0, c^{y}_0, c^{x}_1, c^{y}_1, x_1, y_1\right)_{i}^{j},
    \label{equ:vec}
\end{equation}
where $T$ denotes the basic command type, $(x_0, y_0)$ and $(x_1, y_1)$ represent the beginning and the end points, with $(c^{x}_0, c^{y}_0)$ and $(c^{x}_1, c^{y}_1)$ as the control points of each basic command. Then, to establish interconnections among the adjacent commands, we set the end point of $j$-th command $(x_1, y_1)_{i}^{j}$ equal to the beginning point $(x_0, y_0)_{i}^{j+1}$ of the subsequent $(j+1)$-th command in each individual path.

We then decompose all the paths within the SVG $\mathcal{G}$ into distinct basic commands and combine their corresponding vectors into a matrix form:  \vspace{-1em}
\begin{equation}
\begin{aligned}
    f(\mathcal{G}) &= \left(f(P_i)\right)_{i=1}^{N} = \left(\left(f(\mathcal{C}_{i}^{j})\right)_{j=1}^{M_{i}}\right)_{i=1}^{N} \\ 
    & = \begin{pmatrix}
        (\mathcal{K}_{1}^{1}; & \mathcal{K}_{1}^{2}; & \cdots; & \mathcal{K}_{1}^{M_{1}}) \\
        \vdots & \vdots & \ddots & \vdots \\
        (\mathcal{K}_{N}^{1}; & \mathcal{K}_{N}^{2}; & \cdots; & \mathcal{K}_{N}^{M_{N}})
    \end{pmatrix},
\end{aligned}
\label{equ:matrix}
\end{equation}
where ``;'' denotes the stack operation, and each matrix row represents an individual command. Thus, we can obtain a structured SVG matrix $f(\mathcal{G})\in \mathbb{R}^{(\sum_{i=1}^{N}M_{i})\times 9}$ to represent an SVG that contains $\sum_{i=1}^{N}M_{i}$ individual basic commands.

\begin{figure}[t]
    \centering
    \includegraphics[width=0.98\columnwidth]{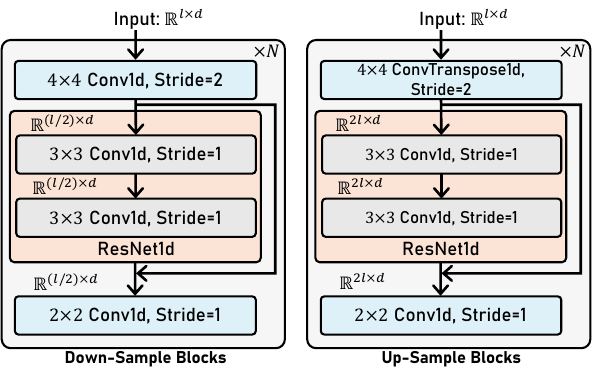}
    \vspace{-1em}
    \caption{Architecture of Down-Sample and Up-Sample Blocks.}
    \label{fig:arch-down-up-sample}
    \vspace{-1em}
\end{figure}

\begin{figure}[t]
    \centering
    \includegraphics[width=0.99\columnwidth]{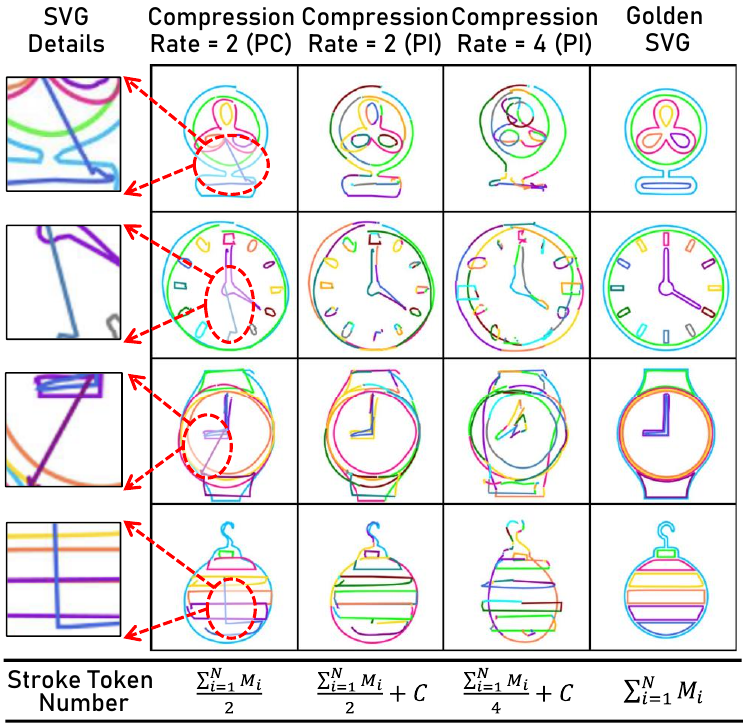}
    \vspace{-0.5em}
    \caption{Analysis of SVG reconstruction, where $C$ is a constant representing the number of inserted $\texttt{<M>}$ command in PI setting. To facilitate clear observation of the SVG composition, we represent each basic command with a distinct color.}
    \label{fig:analysis}
    \vspace{-1em}
\end{figure}

\paragraph{Matrix to Stroke}
After obtaining the SVG matrix $f(\mathcal{G})$, we aim to compress the matrix into discrete stroke tokens via latent representation, with which one can reconstruct the $f(\mathcal{G})$.
As shown in Fig.~\ref{fig:overview-VQ-Stroke}, the VQ-Stroke model is composed of Down-Sample blocks, a Stroke Codebook $\mathcal{B}$, and Up-Sample blocks.
The SVG matrix $f(\mathcal{G})$ is first encoded by the Down-Sample blocks to obtain the compressed representations, which entails increasing the number of representation channels (column of $f(\mathcal{G})$) while concurrently compressing the spatial dimensions (row of $f(\mathcal{G})$) to yield a more compact representation, i.e. compressing the number of commands into $T$ s.t. $T < \sum_{i=1}^{N} M_i$.
Then, the Codebook $\mathcal{B}$ simultaneously conducts $d$ levels of compression with residual vector quantization~\cite{martinez2014stacked}, enabling VQ-Stroke to better model the compressed representations. We depict the detailed architecture of Down-Sample blocks and Up-Sample blocks in Fig~\ref{fig:arch-down-up-sample}, wherein both blocks first utilize a Conv1d or ConvTranspose1d model to compress or expand the features, succeeded by a ResNet1d module and an additional Conv1d module for feature extraction. It is worth mentioning that a low compression rate allows the VQ-Stroke to learn the fine details of SVGs~(the first and second columns), while more aggressive compression~(the third column) enables the VQ-Stroke to capture the overall contours of the SVGs, 
As illustrated in Fig.\ref{fig:analysis}, a low compression rate allows the VQ-Stroke to learn the fine details of SVGs~(the first and second columns), while more aggressive compression~(the third column) enables the VQ-Stroke to capture the overall contours of the SVGs. We have more discussion in Sec.~\ref{subsec:quantitative}.
Finally, the Down-Sample blocks reconstruct the SVG latent representation output from the Codebook $\mathcal{B}$.

To train such a network, we follow~\citeauthor{dhariwal2020jukebox} to calculate the commitment loss, codebook loss, and reconstruction loss to jointly update the VQ-Stroke in Equ.~\ref{equ:VQ-Stroke_loss}: 
\begin{equation}
\begin{aligned}
    & \ell_{VQ-Stroke} = \alpha\left(\ell_{codebook} + \ell_{commit}\right) + \ell_{recon} \\
    &= \alpha \left( \mid\mid\mathcal{Z} - \mathrm{sg}[\Tilde{\mathcal{Z}}]\mid\mid_{2}^{2} + \mid\mid\mathrm{sg}[\mathcal{Z}] - \Tilde{\mathcal{Z}}\mid\mid_{2}^{2} \right) \\
    &+\mathrm{MSE}(\widetilde{f(\mathcal{G})}, f(\mathcal{G})),
\end{aligned}
\label{equ:VQ-Stroke_loss}
\end{equation}
where $\alpha$ is the hyper-parameter, $\mathcal{Z}$ is the compressed latent output from down-sample blocks, $\Tilde{\mathcal{Z}}$ is the latent looked up from codebook $\mathcal{B}$, and $\mathrm{sg[\cdot]}$ is the gradient clipping operation. Besides, we pre-normalize the input data into the $[-1, 1]$ range to stabilize the training process. 

\subsubsection{Encoder-Decoder-based LLM}
We employ an Encoder-Decoder LLM~(EDM) to predict the stroke tokens obtained from the codebook $\mathcal{B}$. 
Considering LLM's inherent textual instruction capability, we freeze the EDM encoder to leverage its inherited textual knowledge. Subsequently, we fine-tune the EDM decoder to learn the stroke token prediction task. Due to the discrepancy between the vocabulary of stroke tokens and the original LLM's vocabulary, we extend EDM with an additional stroke embedding layer and a stroke predictor. Consequently, given the trainable model parameters $\theta$ and the textual prompt $\boldsymbol{K}$, we maximize the log probability $\mathrm{argmax}_{\theta}\prod_{i=1}^{T}P(t_{i}\mid t_{<i}, \boldsymbol{K})$ with the cross-entropy loss.

\subsubsection{SVG Fixer}
A critical issue arises in the generation results from both SDM and EDM, as they fail to guarantee Equ.~\ref{equ:vec} due to the discrepancies of the interconnection points among adjacent commands in each individual SVG path, i.e., $(x_1, y_1)_{i}^{j} \neq (x_0, y_0)_{i}^{j+1}$ in $i$-th path. 
To address this issue, we introduce the SVG Fixer~(SF) as a post-processing module for the generated results. It encompasses two strategies: Path Clipping~(PC) and Path Interpolation~(PI). 
Specifically, PC involves the direct substitution of each SVG command's beginning point with the endpoint of adjacent SVG commands: $(x_0, y_0)_{i}^{j+1} := (x_1, y_1)_{i}^{j}$. On the other hand, PI entails the addition of \texttt{M} commands between each pair of adjacent, yet non-interconnected SVG commands to bridge the discrepancy, i.e., if $(x_1, y_1)_{i}^{j} \neq (x_0, y_0)_{i}^{j+1} \implies$ adding an extra command $\left(\texttt{M}, (x_1, y_1)_{i}^{j}, 0, 0, 0, 0, (x_0, y_0)_{i}^{j+1}\right)$ to force the previous command's end point to move to the beginning point of the next adjacent command.
As shown in Fig.~\ref{fig:analysis}, PC can streamline the overall paths of SVGs, making them more succinct, but may lead to some inaccuracies in the details. On the other hand, PI tends to reveal more generated stokes' details, but it may introduce more curves. Each strategy has its own applicable scenarios.
\section{Experiment}
\label{sec:experiment}

\subsection{Experimental Settings}

\begin{table*}[t]
\small
    \centering
    \begin{tabular}{l | c c c | c c c| c}
    \toprule
     \multirow{2}{*}{Methods} & \multicolumn{3}{c|}{Visual Performance} & \multicolumn{3}{c|}{SVG Code Quality} & \multirow{2}{*}{\makecell[l]{Generation \\ Speed~($\downarrow$) \\ (\textit{per} SVG)}} \\
     \cmidrule{2-7}
     & FID~($\downarrow$) & CLIPScore~($\uparrow$) & HPS~($\uparrow$) & \makecell[c]{Recall~($\uparrow$) \\(Stoke Token)} & EDIT~($\downarrow$) & \makecell[c]{Optim / Pred \\ Length~(\textit{Avg})} \\
     \midrule
     SD \& LIVE & 14.236 & 12.908 & 11.210 & 0.028 & - & 160~(32 Path) & $\approx 28.0$ min \\
     VectorFusion & \underline{7.754} & \underline{17.539} & \underline{15.901} & 0.079 & - & 2,048~(128 Path) & $\approx 30.0$ min \\
     Iconshop & 17.828 & 8.402 & 8.234 & \underline{0.114} & \underline{24,792.476} & 993.244 & $\approx$ \underline{63.743} sec \\
     \midrule
     SVGNUWA~(PC) & 6.607 & 17.852 & 16.134 & \bf 0.239 & \bf 9,092.476 & 271.420 & $\approx \textbf{19.128} $ sec \\
     SVGNUWA~(PI) & \bf 6.513 & \bf 17.994 & \bf 16.801 & 0.207 & 12,249.091 & 271.420 & $\approx \textbf{19.128} $ sec \\
    \bottomrule
    \end{tabular}
    \caption{Performance of StrokeNUWA, where ``Optim/Pred Length'' denotes the actual predicted or optimized number of paths.}
    \vspace{-1.5em}
    \label{tab:main_tab}
\end{table*}

\paragraph{Dataset}
We construct the training and evaluation data with FIGR-8-SVG dataset~\cite{clouatre2019figr}, which consists of massive monochromatic~(black-and-white) SVG icons. We pre-process the SVG data by transforming each SVG sample into standardized representations, eliminating the redundant SVG paths, dropping the outer black box, and filtering the data by applying a threshold of 1,024 basic commands in length.
We filter the instance with less than two annotated discrete keywords and apply a template ``\texttt{Generating SVG according to keywords:\{$\cdots$\}}'' to build the text prompt.
After pre-processing, we sample 2,000 instances with varying SVG code lengths as a testing set, 8,000 samples for validation, and apply the remaining 740K samples for training.

\paragraph{Evaluation Metrics}
We evaluate the quality of the generated SVG of VQ-Stroke and StrokeNUWA from various aspects.
For VQ-Stroke, we primarily consider the reconstruction quality and the compression effectiveness. 
We evaluate the reconstruction quality with the Fréchet Inception Distance~(FID)\footnote{Specifically, we obtain the image features of rendered SVG graphics with the CLIP image encoder~\cite{radford2021learning}}~\cite{heusel2017gans} and the CLIPScore~\cite{radford2021learning}.
Given that the generated SVG graphics consist solely of lines, we set the background color to white to mitigate the potential biases for FID and CLIPScore brought by the background~\cite{wu2023iconshop}.
Additionally, we calculate the Edit Score~(EDIT) between the reconstructed SVG code and the Golden SVG code to reflect the fidelity of the reconstructed SVG graphics in replicating fine details.
To reflect the practical compression effectiveness of VQ-Stroke, we calculate the Compression Ratio~(CR) score between the tokenized SVG code and the stroke tokens, i.e., $\mathrm{CR}=\mathrm{Len}(\mathrm{tokenized~SVG~code}) / \mathrm{Len}(\mathrm{stroke~tokens})$.
For StokeNUWA, apart from utilizing the metrics mentioned above, we supplement evaluation with Human Preference Score~(HPS)~\cite{wu2023human} and Recall Score\footnote{We convert all SVGs into the stroke token format, subsequently computing the recall rate. } to reflect the quality of the generated SVG graphics and their degree of overlap with the Golden SVG Code.
Additionally, we also report the time required to generate each SVG and conduct the qualitative evaluation.
\vspace{-0.5em}
\paragraph{Tasks and Baselines}
We evaluate VQ-Stroke and SVGNUWA with the SVG reconstruction and the text-guided SVG generation tasks, respectively.
For VQ-Stroke, considering the absence of works in the field of SVG representation, we focus on comparing the performance of two SF methods, i.e., PI and PC. Additionally, we evaluate the reconstruction performance of two different compression rates, i.e.,  compression rates of 2 and 4.
For SVGNUWA, we compare with the optimization-based methods, including Vector Fusion~\cite{jain2023vectorfusion} and the Stable Diffusion~\cite{rombach2022high} combined with LIVE method~\cite{li2020differentiable}. Given that optimization-based methods are notably time-intensive, i.e., requiring more than 20 minutes to generate a single SVG on one NVIDIA V100 GPU, we randomly sample 500 instances from the testing set for evaluation to ensure a feasible timeframe.
Additionally, we also compare with the LLM-based method Iconshop~\cite{wu2023iconshop}.  We re-implement Iconshop with the same Flan-T5 backbone as in StrokeNUWA and use a T5 tokenizer to encode the numerical values built in Iconshop.
Notably, the primary distinction between Iconshop and StrokeNUWA lies in their approaches to handling visual representation. While Iconshop directly treats SVG code as visual tokens, StrokeNUWA converts SVG code into stroke visual tokens with VQ-Stroke. We set the maximum model length to 1,500 for IconShop to ensure the completeness of the SVG code.

\begin{table}[t]
    \centering
    \resizebox{\columnwidth}{!}{
    \begin{tabular}{l | c c c | c}
    \toprule
     Methods & FID~($\downarrow$) & CLIPScore~($\uparrow$) & EDIT~($\downarrow$) & CR~($\downarrow$)   \\
     \midrule
        SQM (C-2) & - & - & 1,114.791 & 8.549\%\\
        SQM (C-2) + SF~(PC)  &  3.751 & 19.861 & \bf 1,096.313 & 8.786\% \\
        SQM (C-2) + SF~(PI) & \bf 3.518 & \bf 20.290 & 1,315.137 & 13.780\% \\
        SQM (C-4) + SF~(PI)  &  4.943 & 17.192 & 2,100.671 & \bf 6.890\% \\
        \midrule
        Golden SVG & - & - & - & 100\% \\
    \bottomrule
    \end{tabular}}
    \caption{Performance of VQ-Stroke on SVG reconstruction task, where C-2 and C-4 denote the Compression Rate 2 and 4.}
    \vspace{-1em}
    \label{tab:svg_reconstruction}
\end{table}


\begin{figure}[t]
    \centering    \includegraphics[width=0.99\columnwidth]{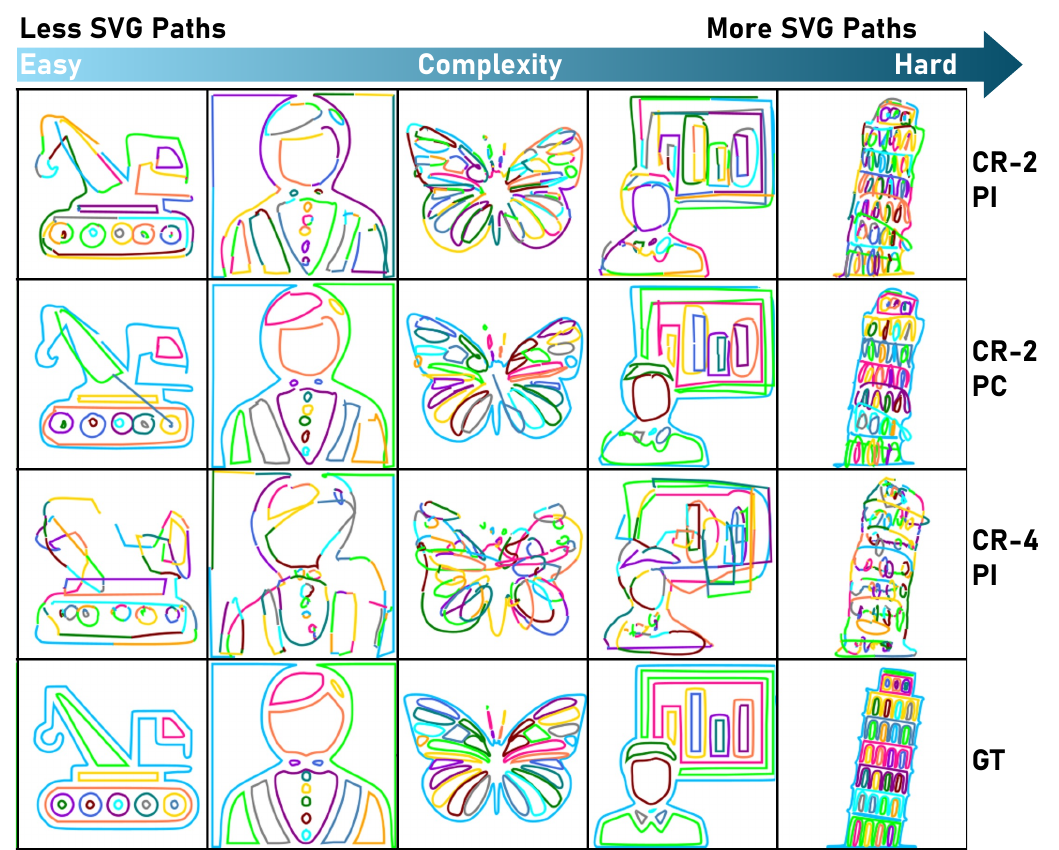}
    \vspace{-0.5em}
    \caption{Reconstruction cases generated by VQ-Stroke, difficulty~(reflected by path numbers) increases from left to right.}
    \vspace{-0.5em}          
    \label{fig:NUWA-stroke-qu}
\end{figure}

\vspace{-1em}
\paragraph{Implementation Details}
For VQ-Stroke, we set the depth of the residual vector quantization $d$ to 2, corresponding to compression rates of 2 and 4.
Then, we set the codebook size $\mid\mathcal{B}\mid$ as 4096, with each code corresponding to a latent representation of 512 dimensions. 
We set $\alpha=1$ in Equ.~\ref{equ:VQ-Stroke_loss} during the training process.
For EDM, we utilize the 3B Flan-T5 model~\cite{chung2022scaling} as the backbone.
We utilize DeepSpeed Library~\citep{rajbhandari2020zero} to implement models on 64 NVIDIA V100 GPUs and set the maximum model length as 512.

\begin{figure}[t]
    \centering
    \includegraphics[width=0.99\columnwidth]{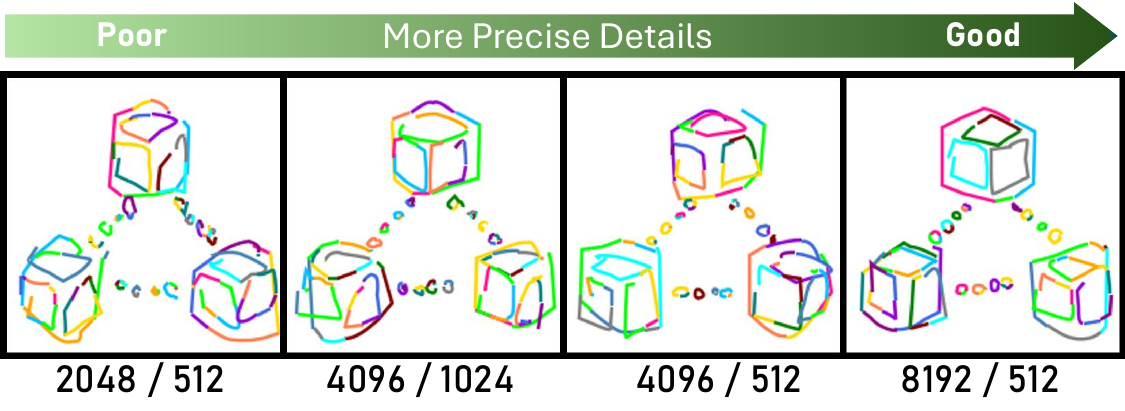}
    \vspace{-1em}
    \caption{Reconstruction performance of difference VQ-Strokes.}
    \label{fig:reconstruction_vq_stroke}
    \vspace{-1em}
\end{figure}

\subsection{Quantitative Evaluation}
\label{subsec:quantitative}

\paragraph{VQ-Stroke}
We report the reconstruction quality of VQ-Stroke in Tab.~\ref{tab:svg_reconstruction}. without SF, VQ-Stroke fails to generate results that conform to SVG syntax. After equipping VQ-Stroke with SF, PI facilitates a more faithful approximation of the original SVG graphics by achieving the lowest FID score and demonstrating a higher concordance with the given text prompts, as evidenced by the lowest CLIP score.
In contrast, the PC method yields better alignment results with the original SVG code as it achieves the lowest EDIT score.
Utilizing compression level 2~(C-4), VQ-Stroke attains a notable Compression Ratio~(CR) of 6.9\%, maintaining performance on par with that of C-2 as evidenced by comparable CLIPScore and FID. This suggests that VQ-Stroke preserves the semantic integrity of the original SVG graphics despite the substantial path compression.

\paragraph{StrokeNUWA}
As illustrated in Table~\ref{tab:main_tab}, StrokeNUWA outperforms other methods by achieving superior results. Specifically, in terms of visual performance, StrokeNUWA is capable of generating graphics that more closely resemble the Golden SVG—evidenced by the lowest FID score~(6.513) and the highest HPS~(16.801). This indicates that our Stroke Tokens offer greater compatibility with the LLMs than the vanilla approach~(Iconshop). Moreover, StrokeNUWA has attained the highest CLIPScore~(17.994), surpassing even Optimization-based methods. This suggests that StrokeTokens encapsulates visual semantics effectively. In terms of the quality of the SVG Code and the efficiency of generation, the Stroke Token not only aligns closely with the Golden standard but also markedly enhances the generation speed, i.e., around 19 seconds of StrokeNUWA V.S. around 30 minutes of Optimization-based method LIVE. This underscores the impressive compressive capabilities of the Stroke token on the original SVG Code, demonstrating both its efficiency and the quality of compression.

\begin{figure}[t]
    \centering
    \subfigure[Comparison between StrokeNUWA and other baselines.]{
    \includegraphics[width=0.95\columnwidth]{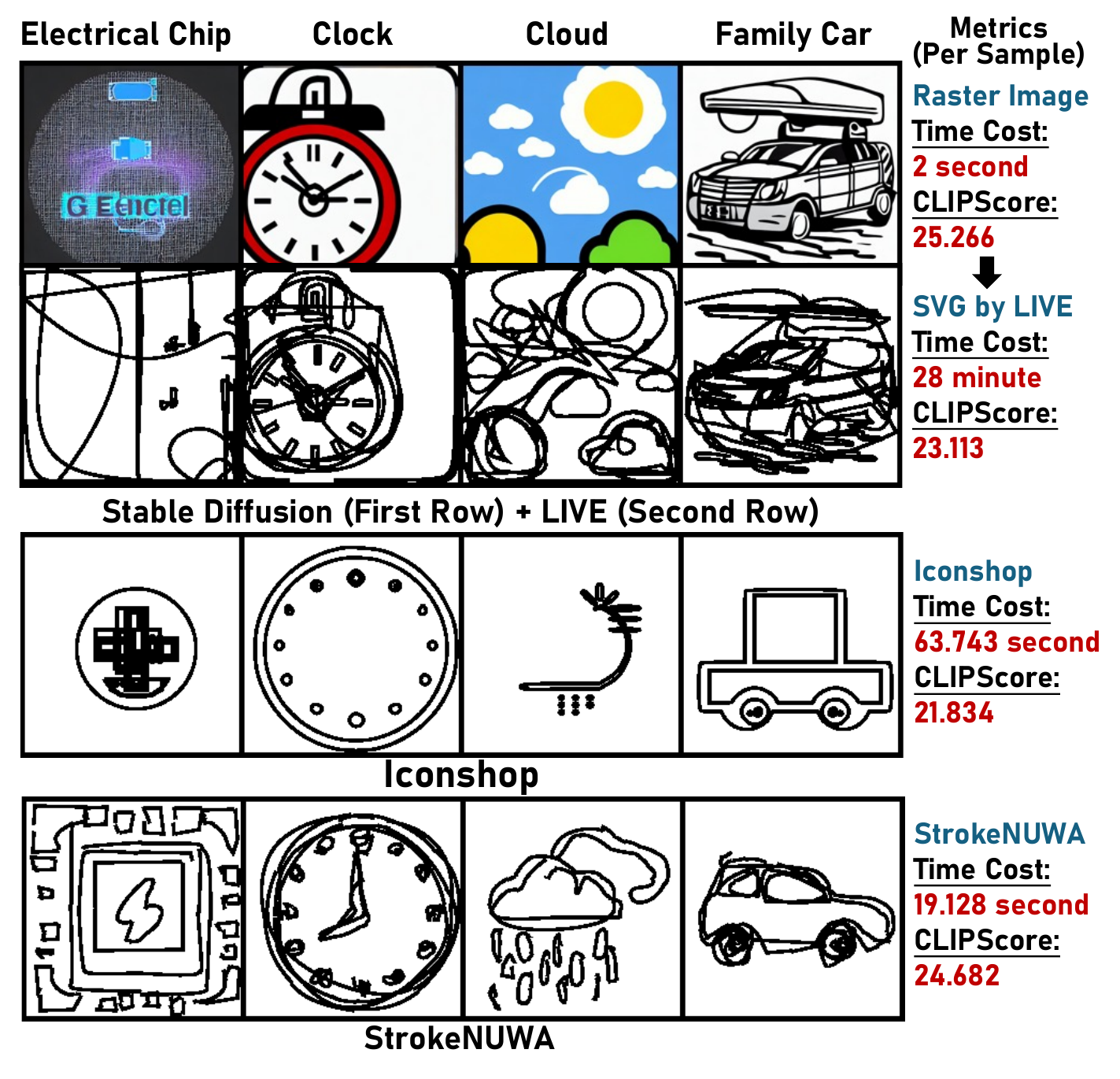}
    \label{fig:vq-stroke-qu}}
    \subfigure[Cases generated by GPT-4-Turbo with same keywords. As GPT-4 is not open-source, we cannot get the generation time.]{
    \includegraphics[width=0.95\columnwidth]{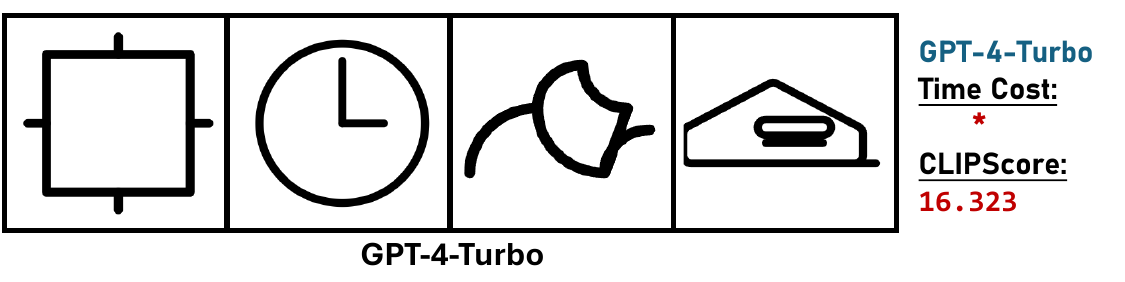}
    \label{fig:gpt4}}
    \vspace{-0.5em}
    \caption{Sampled cases from different models in SVG generation task, where the CLIPScore is the average score calculated across four generated cases for each method.}
    \vspace{-1em}
\end{figure}

\begin{figure}[t]
    \centering
    \includegraphics[width=\columnwidth]{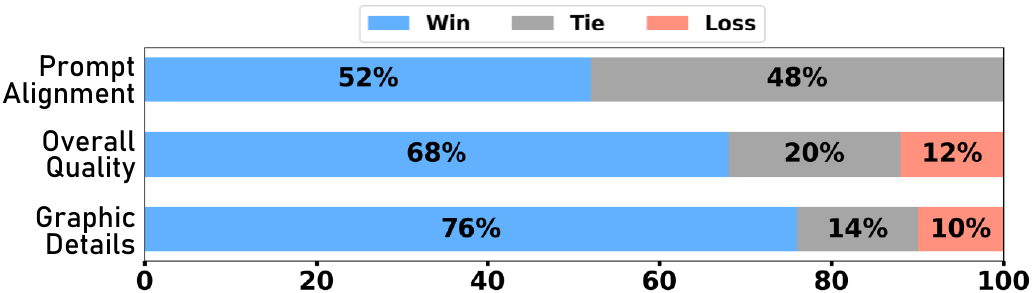}
    \vspace{-1em}
    \caption{Human evaluation between StrokeNUWA and LLM-based method---Iconshop.}
    \label{fig:human_eval}
    \vspace{-1em}
\end{figure}

\subsection{Qualitative Evaluation}
\label{subsec:qualitativa_eval}
\paragraph{Case Study}
We show the reconstruction results of VQ-Stroke with varying complexity levels in Fig.~\ref{fig:reconstruction_vq_stroke} and present a qualitative comparison between StrokeNUWA and other baselines in Fig.~\ref{fig:vq-stroke-qu}.
It is impressive that VQ-Stroke can reconstruct complex SVGs with a limit of only 4,096 codebook size. Then, at the Compression Rate of 2~(CR-2), VQ-Stroke successfully outlines the edge of objects within the graphics, demonstrating that stroke tokens can be highly compressed with a dense representation and inherently incorporate semantic segmentation, which is essential for retaining visual semantics.
Regarding the comparison of StrokeNUWA, we note that employing LLM-based generation methods can result in incomplete SVGs~(Iconshop). This is attributed to the excessive SVG code lengths and LLMs struggling to capture the key information embedded within SVG graphics. However, the use of stroke tokens can mitigate these issues by compressing the paths and being compatible with LLMs. Furthermore, we find that the performance of the optimization-based method heavily relies on the outputs generated by the stable diffusion model, which is subject to the limitations of grid tokens mentioned in Sec.~\ref{sec:intro}, e.g., it is hard to capture the visual semantics and tends to generate extra visual information that is not aligned with the text prompt. Besides, the optimization process is extremely slow. In contrast, StrokeNUWA, which utilizes stroke tokens, inherently contains visual semantic segmentation. As a result, the content generated is more aligned with the textual semantics, providing a more coherent and semantically accurate graphic.
\vspace{-1em}
\paragraph{Human Evaluation}
Furthermore, we conduct a human evaluation to compare the generated SVG outputs from StrokeNUWA with those produced by the LLM-based method, Iconshop. 
We select 50 different textual prompts and guide the model to generate corresponding SVGs for evaluation.
As depicted in Figure~\ref{fig:human_eval}, our comparison is founded on three criteria: Prompt Alignment~(consistency between the generated result and the text prompt), Overall Quality~(the general caliber of SVGs), and Graphic Details~(intricacies such as curves). 
We observe that StrokeNUWA, compared to Iconshop, which regards code as visual representation, not only yields more complete content~(better Overall Quality) but produces results more closely aligned with the textual prompts~(better Prompt Alignment)\footnote{The main reason for low Prompt Alignment in Iconshop is also due to the incompleteness of the generated SVGs.}. Given that stroke tokens compress the details of SVG, it is natural that StrokeNUWA excels in generating Graphic Details.

\section{Ablation Study}
\label{sec:ablation}
\vspace{-0.5em}
\subsection{Analysis of VQ-Stroke Model Architecture}
\label{VQ-Stroke-analysis}
\begin{table}[t]
    \centering
    \begin{tabular}{c c | c c c}
    \toprule
    \multicolumn{2}{c | }{Settings} & \multirow{2}{*}{FID~($\downarrow$)} & \multirow{2}{*}{CLIPScore~($\uparrow$)} & \multirow{2}{*}{EDIT~($\downarrow$)} \\
    \cmidrule{1-2}
    $\mid\mathcal{B}\mid$ & Dim \\  
    \midrule
     2048 & 512 & 5.702 & 19.365 & 2,323.810 \\
     4096 & 512 & 3.518 & 20.290 & 1,315.137 \\
     4096 & 1024 & 3.901 & 20.159 & 1,793.008 \\
     8192 & 512 & \bf 2.639 & \bf 21.014 & \bf 907.106 \\
    \bottomrule
    \end{tabular}
    \caption{Comparison among different VQ-Stroke Settings.}
    \vspace{-1em}
    \label{tab:vq-stroke-analysis}
\end{table}
To investigate the impact of VQ-Stroke architecture configurations on the stroke token performance, we experiment with different codebook sizes $\mid \mathcal{B}\mid$ and codebook dimension $\mathrm{Dim}$. As shown in Tab.~\ref{tab:vq-stroke-analysis}, we can observe that by increasing the codebook size while simultaneously reducing the dimension of each stroke token, the VQ-Stroke achieves superior performance across multiple metrics. We sample a set of reconstruction cases to showcase the trend of changes in Fig.~\ref{fig:reconstruction_vq_stroke}, which indicates that, with a larger codebook size and smaller dimension, the VQ-Stroke can delineate details with greater accuracy, e.g., straighter lines.

\vspace{-0.5em}
\subsection{Comparison with GPT-4}
\label{subsec:decoding-strategy}

We compare the generation results with GPT-4~\cite{achiam2023gpt} by employing the following template to guide GPT-4 in producing the corresponding SVG code:
\texttt{Generate SVG codes in icon style based on keywords:\{KEYWORDS\}}. We show the rendered SVGs in Fig.~\ref{fig:gpt4}, where we can observe that GPT-4 can only generate simple SVGs, which is consistent with LLM-based methods. Moreover, GPT-4 often yields SVGs that are incongruent with the associated text.
\vspace{-0.5em}
\section{Conclusion and Future Work}
This paper presents StrokeNUWA, a pioneering study that explores a superior visual representation---``stroke'' tokens, as an alternative method for expressing images through vector graphics. Stroke tokens not only preserve the semantic integrity of the images but are also conducive to processing by LLMs. Moreover, strokes in vector graphics can be highly compressed. Experiments indicate that, equipped with stroke tokens, LLMs can achieve superior results across various metrics in the SVG synthesis task.
This paper showcases the tremendous potential of stroke token representation in the field of vector graphic synthesis. Moving forward, we aim to continue improving the quality of stroke tokens through advanced visual tokenization methods tailored for LLMs. In addition, we intend to generalize stroke token utilization to a broader range of tasks~(SVG Understanding), domains~(3D), and the generation of SVGs for images sourced from the real world.

\section*{Impact Statement}
The implications of this work are manifold, potentially revolutionizing the visual synthesis from another format of image, vector graphics. As stroke tokens refine the interplay between visual representation and LLMs, future advancements in visual tokenization techniques designed for LLMs are anticipated. Moving forward, the community can extend stroke token application into wider tasks and domains, including SVG comprehension and open-domain SVG synthesis for images from the real world.
As we pioneer this nascent field, we are conscious of the profound societal impact that such advancements in machine learning and graphical representations hold. The capabilities for automated graphic design, scalable vector graphics production, and enhanced digital artistry foreshadow considerable shifts in industries reliant on visual content. By forging new pathways for artistic expression and visual communication, our work stands to not only contribute to the scientific community but also to catalyze transformations in creative, technological, and educational sectors.
We recognize the importance of our work and our responsibility to ensure that our contributions to the field are conducted ethically, aiming to benefit society as a whole, democratize the visual landscape, and enrich it through responsible and judicious innovation.

\bibliography{main}
\bibliographystyle{icml2024}

\end{document}


\twocolumn[
\icmltitle{SVGNUWA: Harnessing LLMs for Latent Text-Guided SVG Creation (Appendix)}


\icmlsetsymbol{equal}{*}

\begin{icmlauthorlist}
\icmlauthor{Firstname1 Lastname1}{equal,yyy}
\icmlauthor{Firstname2 Lastname2}{equal,yyy,comp}
\icmlauthor{Firstname3 Lastname3}{comp}
\icmlauthor{Firstname4 Lastname4}{sch}
\icmlauthor{Firstname5 Lastname5}{yyy}
\icmlauthor{Firstname6 Lastname6}{sch,yyy,comp}
\icmlauthor{Firstname7 Lastname7}{comp}
\icmlauthor{Firstname8 Lastname8}{sch}
\icmlauthor{Firstname8 Lastname8}{yyy,comp}
\end{icmlauthorlist}

\icmlaffiliation{yyy}{Department of XXX, University of YYY, Location, Country}
\icmlaffiliation{comp}{Company Name, Location, Country}
\icmlaffiliation{sch}{School of ZZZ, Institute of WWW, Location, Country}

\icmlcorrespondingauthor{Firstname1 Lastname1}{first1.last1@xxx.edu}
\icmlcorrespondingauthor{Firstname2 Lastname2}{first2.last2@www.uk}

\icmlkeywords{Machine Learning, ICML}

\vskip 0.3in
]



\printAffiliationsAndNotice{\icmlEqualContribution} 

\nocite{langley00}

\newpage
\appendix
\section{EDM Training Phrase}

\begin{figure}[t]
    \centering
    \includegraphics[width=\columnwidth]{icml2024/figure/EDM.pdf}
    \caption{Training phrase of EDM.}
    \label{fig:edm}
    \vspace{-1em}
\end{figure}

\bibliography{example_paper}
\bibliographystyle{icml2024}